\pdfoutput=1
\documentclass[11pt,a4paper]{article}

\usepackage{times}
\usepackage{latexsym}

\usepackage[acceptedWithA]{tacl2021v1}

\usepackage[T1]{fontenc}
\usepackage[utf8]{inputenc}
\usepackage{microtype}
\usepackage{subcaption}
\usepackage{graphicx}
\usepackage{booktabs}
\usepackage{tabularx}
\usepackage{placeins}
\usepackage[capitalise]{cleveref}
\usepackage{stix}
\usepackage{ragged2e}

\usepackage[symbol]{footmisc}

\definecolor{light-green}{RGB}{118, 207, 180}
\definecolor{yes-answer}{HTML}{ffdc85}
\definecolor{no-answer}{HTML}{3a6a00}
\definecolor{not-sure-answer}{HTML}{a3b18a}

\title{Surveying (Dis)Parities and Concerns of Compute Hungry NLP Research}

\author{
    Ji-Ung Lee$^{1,2}$,
    Haritz Puerto$^{1,2}$,
    Betty van Aken$^{3}$, % still waiting for an OK from Betty
    Yuki Arase$^{4}$, \\
    \bf 
    Jessica Zosa Forde$^{5}$,
    Leon Derczynski$^{6,7}$,
    Andreas Rücklé$^{10,\ddagger}$, \\
    \bf
    Iryna Gurevych$^{1,2}$,
    Roy Schwartz$^{8}$,
    Emma Strubell$^{9,11}$,
    Jesse Dodge$^{11}$
    \\
    $^{1}$Technical University of Darmstadt,
    $^{2}$Hessian AI, 
    $^{3}$Berliner Hochschule für Technik,  \\
    $^{4}$Osaka University,
    $^{5}$Brown University,
    $^{6}$University of Washington, 
    $^{7}$IT University of Copenhagen, \\
    $^{8}$The Hebrew University of Jerusalem,
    $^{9}$Carnegie Mellon University,
    $^{10}$Amazon,
    $^{11}$Allen Institute for AI
}
  
\begin{document}
\maketitle
\begin{abstract}
Many recent improvements in NLP stem from the development and use of large pre-trained language models (PLMs) with billions of parameters.
Large model sizes makes computational cost one of the main limiting factors for training and evaluating such models; and has raised severe concerns about the sustainability, reproducibility, and inclusiveness for researching PLMs. 
These concerns are often based on personal experiences and observations. 
However, there had not been any large-scale surveys that investigate them.
In this work, we provide a first attempt to quantify these concerns regarding three topics, namely, \textit{environmental impact}, \textit{equity}, and \textit{impact on peer reviewing}. 
By conducting a survey with 312 participants from the NLP community, we capture existing (dis)parities between different and within groups with respect to seniority, academia, and industry; and their impact on the peer reviewing process.
For each topic, we provide an analysis and devise recommendations to mitigate found disparities, some of which already successfully implemented.
Finally, we discuss additional concerns raised by many participants in free-text responses.
\end{abstract}

% This is required
\renewcommand*{\thefootnote}{\fnsymbol{footnote}}
\footnotetext[3]{This work does not relate to AR's position at Amazon.}
\renewcommand*{\thefootnote}{\arabic{footnote}}

\section{Introduction}
 Recent advances in hardware and algorithms have transformed the field of NLP. 
Whereas NLP practitioners and researchers used to be able to develop and use cutting-edge NLP technology on relatively affordable hardware such as a laptop or a commodity server, modern state-of-the-art approaches have evolved to require more substantial computational power, typically achieved by specialized tensor processing hardware such as a GPU or TPU. 
This shift has raised at least two concerns among members of the NLP community~\citep{strubell-etal-2019-energy,schwartz2020green,ACL-policy-doc-2021} and the AI community~\citep{patterson2022carbon,wu2022sustainable}: 
(1) Understanding and mitigating the environmental cost of NLP research and use, in terms of greenhouse gas (GHG) emissions, and 
(2) equity of access: the extent to which increasing computational requirements restricts who has access to develop and use modern NLP. 

\begin{figure}
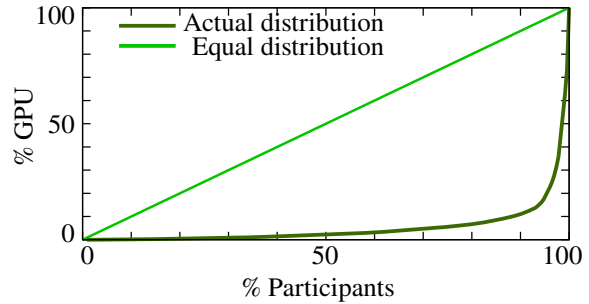

    \centering
    \include{final_plots/q4_percentage}
    \vspace{-3em}
    \caption{Distribution of available GPUs across our participants (in \%). As can be seen, 87.8\% of our survey participants have access to less than 10\% of the total number of GPUs.} 
    \label{fig:q4_dist}
\end{figure}

% Figure for section 2

In response to these concerns, we formed a working group within ACL with the goal of better understanding the challenges surrounding efficient NLP and establishing policies to address them. 
In order to quantify views and impacts in the NLP community related to these concerns, we conducted a survey of the ACL community in July 2021, the results of which we report here. 
Besides concerns about the (1) environmental impact and (2) equity, we further solicit answers about their (3) impact on the whole peer reviewing process, as this is an important matter for inclusiveness. 
Overall, we elicited 312 responses from a distributed range of junior and senior researchers hailing from industry and academia.
Some of our key findings include:

\begin{itemize}
    \item More than 50\% of the survey participants are moderately or very concerned about the environmental footprint of NLP research; mostly with respect to \textit{training} and \textit{model selection}.
    \item Overall, $\sim$62\% of our respondents have access to less than eight GPUs and moreover, over 90\% have access to less than 10\% of the total GPU power (\cref{fig:q4_dist}). As a frame of reference, recent work \citep{izsak2021} showed that a clever set of techniques can be used to train BERT in 24 hours on 8 GPUs, and it takes about 7 minutes to fine-tune a RoBERTa$_\mathrm{LARGE}$ model on the MNLI natural language inference dataset (about 400k training sentences) on one GPU (GTX 2080 Ti) to an accuracy of 85\% \citep{zhou-etal-2021-hulk}. 
    \item A majority (76\%) our respondents believe that it would be beneficial to have smaller versions of pre-trained models released together with larger ones. In fact, 33\% of our free-text respondents emphasised the importance of sharing artifacts (such as code, models, training logs, etc.). 
    \item The group that suffers most from lack of resources are students, who struggle to reproduce previous results when compared to researchers from large industry. 
    \item While we find disparities between different groups---especially regarding the job sector---our analysis shows that most of them are not \textit{statistically significant}.
    Instead, we find outliers across all groups showing that there exist disparities within. 
    We find no evidence in our survey responses that ``industry'' has access to significantly more compute power than ``academia''. 
    Instead, this mostly seems to be the case for very few extreme outliers (6\%). 
\end{itemize}

With this survey, we hope to provide a more solid foundation to back-up the ongoing discussion in the community and for devising concrete actions to make research more inclusive.

\section{Survey Description}

The survey was open over a period of 17 days, from Monday, July 12, 2021 to Thursday, July 29, 2021. 
It was conducted via Microsoft Forms and distributed across the *CL community by mass mailing to ACL membership, and shared on Twitter. 
During that time, we collected 312 responses.
The creation of the survey indicated that, ``input will remain anonymous and the responses will also be summarized in aggregate form''.\footnote{\url{https://www.aclweb.org/portal/content/efficient-nlp-survey}} 
Therefore the data will be made available on request with a statement of intended purpose, due to privacy and ethical restrictions.

\subsection{Questionnaire}\label{sec:questions}
The questions were divided into four categories. 
First, we collected some general information about our participants, like their current position and seniority (\S\ref{sec:demographics}).
Second, we asked our participants about their concerns regarding the environmental impact of NLP experiments (\S\ref{sec:env_concerns}) and their access to computational resources (\S\ref{sec:equity}).
Finally, we asked about the impact of compute-intensive experiments on the reviewing process as well as about specific measurements to alleviate them (\S\ref{sec:reviewing}).

To keep a low-effort for our participants, we crafted most of the ($Q$)uestions as simple yes/no/unsure questions. 
For subjects that require a more fine-grained analysis (e.g., environmental concerns) we used five point scales (either numeric or text-based). 
Overall, we asked a total of 19 questions from which 15 were multiple-choice questions (13 with a single answer possibility and two with multiple possible answers).
$Q$4 (available compute resources)  and $Q$11 (number of times reviewers asked for expensive experiments) required a numeric answer.
Finally, $Q$9, $Q18$, and $Q$19 allowed free text answers.
Participants were asked to provide answers to 13 questions, while six questions ($Q4$, $Q9$, $Q11$, $Q12$, $Q14$, $Q19$) were optional.
All questions are provided in \cref{table:questions}.

\subsection{Demographic Overview}\label{sec:demographics}
\begin{figure*}[!ht]
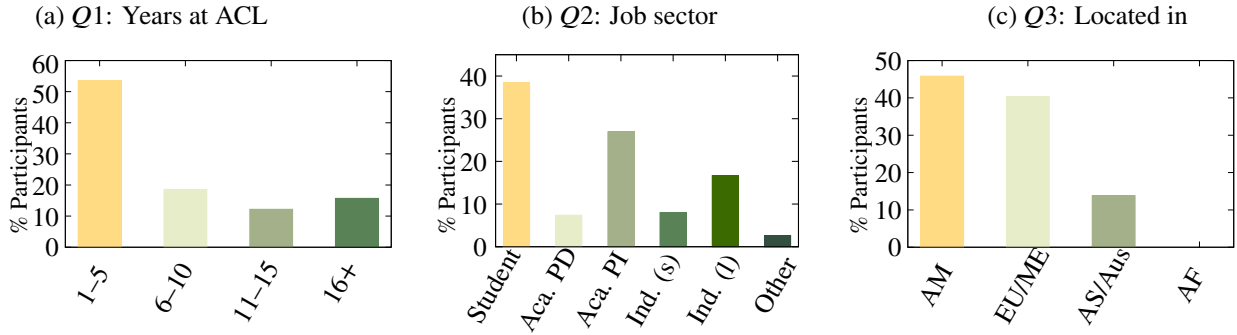

    \begin{subfigure}[b]{0.23\textwidth}
        \centering
        \caption{$Q$1: Years at ACL} 
        \vspace{-4em}
        \include{final_plots/q1_bars.tex}
        \label{fig:q1}
    \end{subfigure}
    \hfill 
    \begin{subfigure}[b]{0.23\textwidth}
        \centering
        \caption{$Q$2: Job sector} 
        \vspace{-4em}

        \include{final_plots/q2_bars.tex}
        \label{fig:q2}
    \end{subfigure}
    \hfill
    \begin{subfigure}[b]{0.23\textwidth}
        \centering
        \caption{$Q$3: Located in} 
        \vspace{-4em}

        \include{final_plots/q3_bars.tex}
        \label{fig:q3}
    \end{subfigure}
    \caption{Demographic statistics: (a) describes the seniority, (b) the job sector, and (c) the geographic location of our participants (in \%).}
    \label{fig:demographics}
\end{figure*}

In our first three questions, we asked the participants about their seniority, job sector, and geographic location (\cref{fig:demographics}). 

\paragraph{Seniority.}
We asked our participants about the number of active years in the *CL community as an author, reviewer, or in a related role ($Q$1).
Overall, a little over half (53.5\%) indicated they were junior members of the community, while the remainder were fairly evenly split across mid- and late-career.

\paragraph{Job Sector.}
We further asked our participants about the current position they are holding ($Q$2).
Possible responses were student, academic post-doc (Aca.~PD), researcher from small (s) and large (l) 

\begin{table*}[t] 
\begin{center}
\begin{small}
\scalebox{0.88}{
\begin{tabularx}{\linewidth}{ X }
\toprule
\textbf{Demographics} \\\midrule
\textbf{Q1. Years active.} How many years have you been active in the ACL community (as an author/reviewer/area chair/etc.)? \textbf{Answer}: [1-5], [6-10], [11-15], [16+]. \\
\textbf{Q2. Current Role.} \textbf{Answer}: Student, Academic Postdoc, Academic PI, Researcher in large industry, Research in small industry, other.           \\
\textbf{Q3. Geographic Location.} \textbf{Answer}:    Americas, Europe/Middle East, Africa, Asia/Oceania.  \\ 
\midrule
\textbf{Equity} \\
\midrule 
\textbf{Q4. Available compute resources}. Please provide a rough estimate of the average number of GPUs or equivalent accelerators that are available to you (for students / researchers) or to each researcher in your lab/group (for PIs / managers). If you cannot
quantify the amount of compute resources, leave this field empty.
\textbf{Answer:} Numeric response (optional). \\
\textbf{Q5. Unable to run experiments}. In the last year, have you been unable to run experiments important for one of your projects\\ due to lack of computational resources? \textbf{Answer}: Yes, No, Unsure. \\
\textbf{Q6. More resources would make your work more valuable}. How often do you feel like your work would have been valued more by the community (e.g., accepted instead of rejected to some venue) if you had access to more computational resources? \textbf{Answer}: Five point scale.\\
\midrule
\textbf{Environmental Concern} \\
\midrule 
\textbf{Q7. Concern about environmental footprint.} How concerned are you by the environmental footprint of the field of NLP? \textbf{Answer}: Five point scale. \\
\textbf{Q8. Most pressing factor.} Which of the following do you feel is the most pressing factor with respect to the environmental impact of NLP? \textbf{Answer}: Choose all that apply: Training, Inference, Model selection, None, Other). \\
\textbf{Q9. Why?} Optionally explain the reasons for your choices above. \textbf{Answer}: Free text (optional). \\
\midrule
\textbf{Reviewing Process} \\
\midrule 
\textbf{Q10. Did reviewers ask for too expensive experiments?} In the past 3 years, have you received feedback from reviewers who requested experiments that were too expensive for your budget for a particular paper? \textbf{Answer:} Yes, No, Not sure.\\
\textbf{Q11. If yes, how many times?} \\
\textbf{Q12. Was the critique justified?} If yes, do you feel the critique was justified? I.e., that the main scientific claims in your paper (e.g., that your approach was better than some baseline) were not sufficiently supported by the original,\\ smaller-budget experiments? \textbf{Answer:} Yes, No, Not sure.\\
\textbf{Q13. Lack of resources prevents reproduction of previous results.} How often do you find yourself unsuccessful in reproducing a previous result due to lack of computational resources? \textbf{Answer}: Never, Rarely, Sometimes, Often, Always. \\
\textbf{Q14. Efficiency Track}. If you have work on efficient methods and/or enhanced reporting, would you consider submitting it to a dedicated track? \textbf{Answer}: Yes, No, N/A. \\
\textbf{Q15. Justify allocation of budget for experiments}. As a reader, would you prefer authors to be requested to justify the way they allocate their budget to run experiments which adequately support their scientific claims? \textbf{Answer}: Yes, No, Not sure.\\
\textbf{Q16. Reviewers should justify the petition for additional experiments}. As an author, would you prefer it if reviewers took up space in their review to justify their suggestions for additional experiments in terms of the evidence that those additional experiments would provide? I.e., what is currently missing in terms of lack of evidence to support the main claims of the paper, and how the additional experiments would provide evidence for the paper’s research questions? \textbf{Answer}: Yes, No, Not sure.\\
\textbf{Q17. Releasing small versions of pretrained models}. Would your work benefit from smaller versions of pretrained models released alongside larger ones? \textbf{Answer}: Yes, No, Not sure.\\
\textbf{Q18. How to encourage the release of models}. Which of these solutions would you endorse for encouraging the release of trained models? \textbf{Answer}:  Choose all that apply: Best artifact award, Instruct reviewers to reward papers who share/promise to share models, Visible branding of the paper in conference proceedings, None of the above, Other)\\
\midrule
\textbf{Q19. Any other thoughts or suggestions?} \textbf{Answer}: Free text (optional). \\

\bottomrule
\end{tabularx}
}
\end{small}
\end{center}
\caption{List of questions in the survey. Summaries of the questions in bold. Only full questions were shown to the participants.}
\label{table:questions}
\end{table*}

\FloatBarrier

\noindent
industries (Ind.), and academic PI (Aca.~PI).
The largest group of participants were students (38.5\%), followed by academic postdocs and PIs (34.3\%), and industry researchers (24.7\%). 
Eight participants (2.5\%) responded with ``other'' from which seven were affiliated with academia (e.g., lecturers) and one with industry (consultant).
For the fine-grained analysis, we merge each response of ``other'' into the most fitting group in the survey (one student, five academic PIs, one academic post-doc, and one small industry researcher).
For our analysis, we do not merge the academic and industry subgroups, as this may obfuscate existing disparities; e.g., between small and large industry.

\paragraph{Geographic location.}
We further asked respondents to share their geographic location ($Q$3).
Overall, 45.8\% of responses came from the Americas (AM), 40.4\% from Europe (EU) and the Middle East (ME), and 13.8\% from Asia (AS) and Australia (Aus). 
We received no responses from researchers in Africa (AF). 
The heavily skewed responses in terms of geographic location limits the expressiveness of this factor and thus, will not be considered for our analysis.

\subsection{Methodology}
In the following sections, we analyse and discuss the participants' responses with respect to the remaining three categories (\textit{environmental concerns}, \textit{equity}, and \textit{impact on the reviewing process}).
For each section, we first provide an overview of the distribution in the responses and then provide a fine-grained analysis with respect to the \textit{seniority} and \textit{job sector}.
The goal of the fine-grained analysis is to investigate if we can observe any statistically significant differences across different groups.

\paragraph{Statistical tests.}
Due to the explorative nature of our survey, the collected data violates the necessary conditions on homoscedasticity~\citep{levene1960robust} and normality~\citep{shaphiro1965analysis} that are required to conduct an analysis of variances (ANOVA, \citealt{fisher1921probable}). 
Instead, we perform a Kruskal-Wallis test~\citep{kruskal1952use} as an indicator for any statistically significant differences\footnote{This is the case when $H > H^n_0$ with $H^n_0 \sim \chi^2_{n-1}$ for $n$ groups. For $\alpha=0.05$, we get $H^5_0 = 9.488$ (job sector) and $H^4_0 = 7.815$ (seniority) \citep{abramowitz1988handbook}.} and if so, perform pairwise Welch's t-tests~\citep{welch1951comparison} against a Bonferroni corrected $\alpha = \frac{0.05}{m}$ where $m$ is the number of pairwise comparisons; i.e., for $n$ groups, $m=\frac{n\cdot(n-1)}{2}$~\citep{bonferroni1936teoria}. 
This results in corrected $\alpha = 0.008\overline{3}$ for seniority with $n=4$ and $\alpha = 0.005$ for the job sector with $n=5$ (not merging academia and industry sectors).
For the numerical questions ($Q$4 and $Q$11), we further analyze if there exist disparities within each group, using interquartile ranges with $k=1.5$ to detect outliers~\citep{tukey77}.

\section{Environmental Footprint}\label{sec:env_concerns}

We quantified existing concerns about the environmental impact of NLP experiments using a five point Likert scale ($Q$7) and asked our participants to select the most pressing issue in the typical life cycle of an NLP model ($Q$8) between (Train)ing, model (Select)tion, and (Infer)ence.
Participants were allowed to select \textit{all} applicable answers and could select (None) or provide (Other) pressing issues.
They could also provide a textual justification of their answer(s) ($Q$9).

\subsection{Analysis}
\begin{figure*}[!htb]
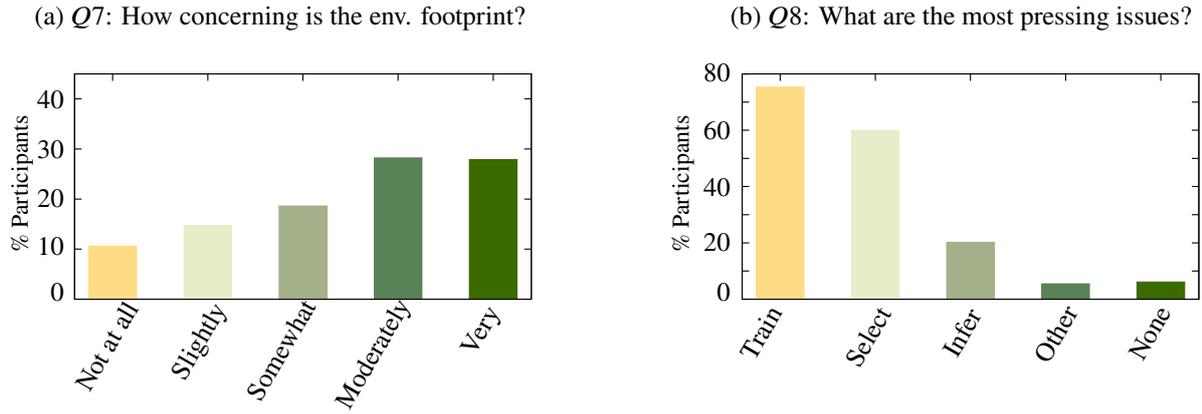

    \begin{subfigure}[b]{0.44\textwidth}
        \centering
        \caption{$Q$7: How concerning is the env. footprint?} 
        \vspace{-2em}
        \include{final_plots/q7_bars.tex}
        \label{fig:q7}
    \end{subfigure}
    \hfill 
    \begin{subfigure}[b]{0.44\textwidth}
        \centering
        \caption{$Q$8: What are the most pressing issues?}
        \vspace{-2em}
        \include{final_plots/q8_bars.tex}
        \label{fig:q8}
    \end{subfigure}
    \caption{Environmental concerns and pressing issues (in \% of participant answers).}\label{fig:env_footprint}
    
\end{figure*}

\Cref{fig:q7} shows that more than 50\% of our participants were moderately (28.2\%) or very (27.9\%) concerned about the environmental footprint of NLP, while around 33\% of them were slightly (14.7\%) or somewhat (18.6\%) concerned.
10.6\% of participants were not concerned at all.
Our participants further agreed that \textit{training} (75.3\%) and \textit{model selection} (59.9\%) are the most pressing issues (\cref{fig:q8}).\footnote{Note that 39.7\% of our participants selected exactly these two as the only pressing factors.}
\textit{Inference} took third place with 20.2\%, while 6.1\% of our participants selected \textit{none}.
The smallest number responses was given for \textit{other} with \textit{hyperparameter tuning} and \textit{travelling} (6 mentions each) being the most frequent ones. 
Also mentioned were \textit{storage consumption}, \textit{hardware}, \textit{expectations about large data experiments}, and \textit{scale}.
Interestingly, many respondents considered inference less pressing than training and model selection.

\begin{figure*}[!htb]
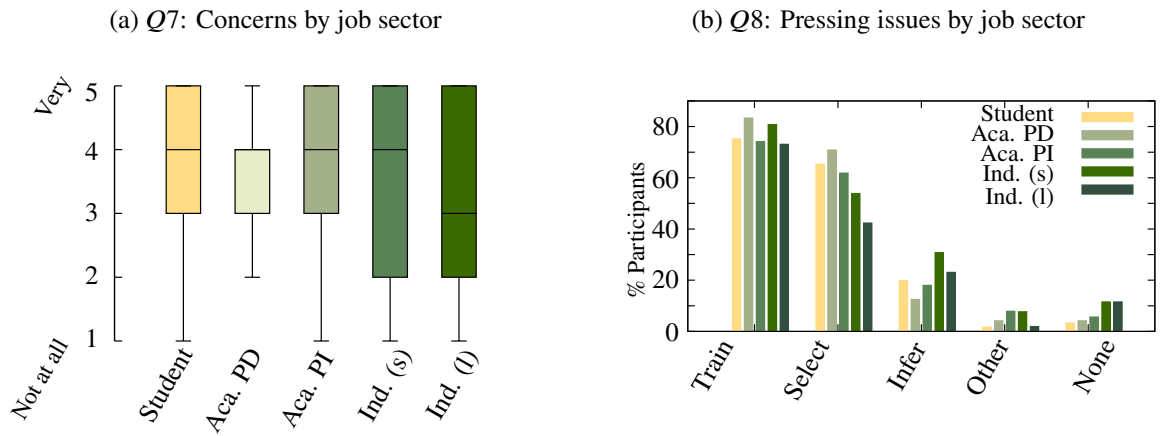

    \begin{subfigure}[b]{0.50\textwidth}
        \caption{$Q$7: Concerns by job sector} 
        \centering
        \vspace{-1em}
        \include{final_plots/q7_job}
        \label{fig:q7-job}
    \end{subfigure}
    \hfill 
    \begin{subfigure}[b]{0.50\textwidth}
        \centering
        \caption{$Q$8: Pressing issues by job sector} 
        \label{fig:q8-job}
        \vspace{-1em}
        \include{final_plots/q8_job}
    \end{subfigure}
    \caption{Concerns and pressing issues, grouped by positions.}
    \label{fig:q7-fine-grained}
\end{figure*}

\paragraph{Job Sector.} Although we do not find significant differences by seniority, we see larger (although not statistically significant) differences when looking at the responses grouped by researchers from different job sectors (\cref{fig:q7-job}).
We find that respondents from the large industry sector were mostly \textit{somewhat} concerned, while the median for all other groups lies at \textit{often} concerned. 
Similarly, we also see larger differences in the most pressing issues between different groups. 
For instance, small and large industries were substantially more concerned with respect to inference and much less concerned with respect to model selection than academia.

\FloatBarrier

\subsection{Discussion and Recommendations}

An analysis of the 81 (26\%) free-text responses ($Q$9) reveals diverse opinions about the environmental impact of NLP and the reasons behind the most pressing factors. 
For instance, among respondents that stated to be \textit{not at all} concerned about NLP's environmental footprint, a majority considered the impact of NLP research on climate change to be negligible compared to other factors. 
Factors mentioned as being more relevant to climate change include air travel (also mentioned twice in the general responses $Q19$), cars, and more cost intensive computations from other areas (of science).
Another argument brought up multiple times in this group of respondents is that the ACL is not the right institution to tackle challenges of climate change. 
Some responses alternatively suggested to push for regulatory changes, since big tech companies might not be affected by decisions made by the ACL.

Regarding the most pressing factors, one of the main arguments provided for inference was that industry spends most time on inference, hence it is the most expensive one.
However, participants also argued that there exist various methods for efficient inference~\citep[see, e.g.,][]{treviso2022efficient}.
Prominent arguments with respect to training and model selection were that the pressure to achieve state-of-the-art performance leads to extensive hyperparameter tuning and that a large variety of models are being trained during research and development (even if just for debugging) without being ever deployed.

\section{Equity}\label{sec:equity}

 \begin{figure}[t]
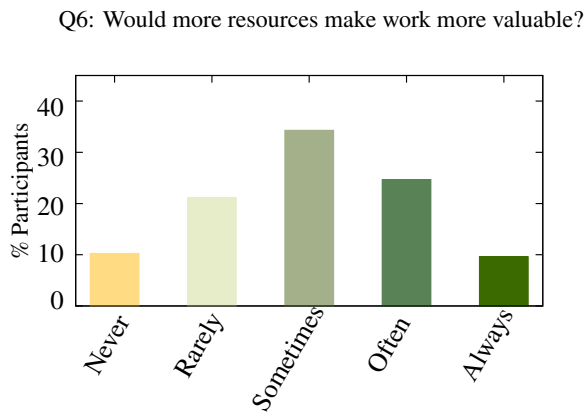
{}
 \include{final_plots/q6_bars}
        \caption{Lack of resources for more valuable work. } 
        \label{fig:q6}
\end{figure}

The (in)equity of the available compute resources across groups (e.g., academia and industry) is an increasingly brought up topic in discussions. 
While the general gist seems to be that many researchers feel excluded by not having access to substantially large compute power (e.g., thousands of GPUs), it often remains unclear whether this is really the case.
One of the main objectives of this survey was therefore to quantify such potential disparities.

\subsection{Analysis}
\begin{figure*}[!htb]
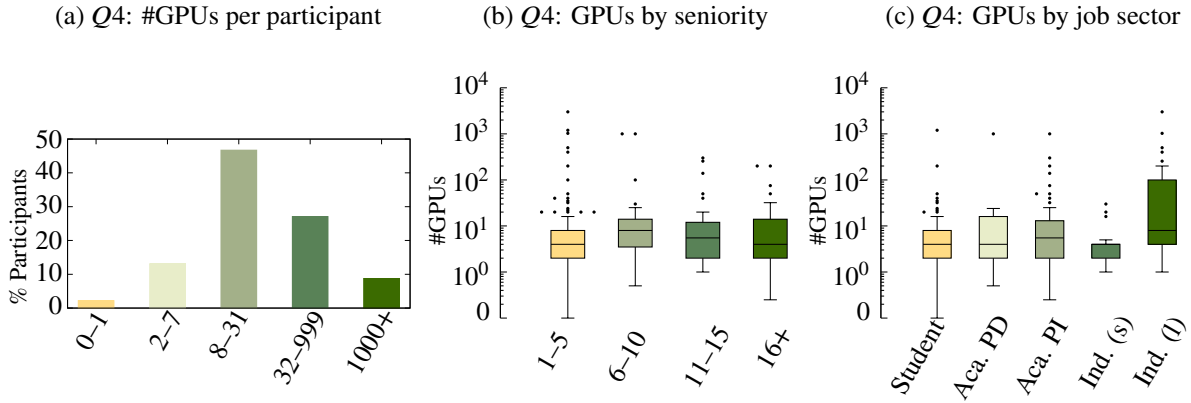

\begin{subfigure}[b]{0.33\textwidth}
        \centering
        \caption{$Q$4: \#GPUs per participant}\label{fig:q4}
        \vspace{-2em}
        \include{final_plots/q4_bars}
    \end{subfigure}
    \begin{subfigure}[b]{0.33\textwidth}
        \centering
        \caption{$Q$4: GPUs by seniority}\label{fig:q4_age}
        \vspace{-2em}
        \include{final_plots/q4_gpu_age}
    \end{subfigure}
    \hfill 
    \begin{subfigure}[b]{0.33\textwidth}
        \centering
        \caption{$Q$4: GPUs by job sector}\label{fig:q4_job}
        \vspace{-2em}
        \include{final_plots/q4_gpu_job}
    \end{subfigure}
    \caption{Distribution of GPUs among participants: (a) overall, (b) by seniority, (c) by job sector.}\label{fig:boxplot_careers_years_gpus}
\end{figure*}

For $Q$4, 229 participants responded (73.4\%) with the number of GPUs they have access to. 
\cref{fig:q4_dist} shows the distribution of the total number of GPUs across our participants (in~\%).
Overall, we observe a high disparity across our participants in terms of access to GPUs. 
For instance, 62\% of the participants had access to less than eight GPUs (\cref{fig:q4}), the number used for training academic BERT~\citep{izsak2021}, and 87.8\% of the participants had access to only 9.7\% of the total number of GPUs. 
15~participants (6.6\%) had access to more than 100 GPUs, up to 3000 GPUs, representing 85.6\% of the total GPU count ($\sim$11.2k). 
An outlier analysis shows that 13.1\% of the respondents had access to a substantially higher number of GPUs (more than 22 GPUs) than the rest.
We further find that 57.4\% of our participants were unable to run experiments due to the lack of computational resources, and 36.2\% had no lack of resources ($Q$5).
Finally, \cref{fig:q6} shows that 31.4\% of our respondents \textit{never} or \textit{rarely} thought that more resources could make their work valuable, while 34.3\% of respondents answered \textit{sometimes}, and 34.3\% answered \textit{often} or \textit{always} ($Q$6).

\paragraph{Job Sector.} As in \S \ref{sec:env_concerns}, our analysis shows no significant differences w.r.t.~the seniority, and we find larger disparities by job sector.
As we can observe in \cref{fig:q4_job}, respondents in industry (large) had access to a higher number of GPUs than industry (small) and academia.
This is one of the few cases where we have to resort to pairwise testing, as the Kruskal-Wallis test indicates that the Null hypothesis cannot be rejected with $H^5 = 16.976 > H_0^5 = 9.488$.
While we do not find significant differences in our pairwise comparisons, there are still substantial differences between Ind.~(l) and Aca.~PI (p-value $=0.0827$), as well as Ind.~(s) and Ind.~(l) (p-value $=0.0850$).
Even though students reported the lowest number of available GPUs, the differences seem less substantial compared to researchers at small industry (p-value $=0.110$). 
Additionally, we find that large industry has the highest percentage of outliers and the largest in-group disparity. 
Interestingly, researchers from small industry seem to have the least issues when running experiments; a stark contrast considering they are among those who reported the fewest GPU resources ($Q$5).
Regarding $Q$6, researchers from large industry responded slightly less often than other groups that their research could be more valuable if they had access to more compute power.

\subsection{Discussion and Recommendations}
While our survey highlights existing disparities, particularly between small industry or academic researchers and large industry, we also find that there exist substantial disparities \textit{within} each group. 
Most surprising might be the general disparity we find across the field, as 87.8\% of our participants had access to less than 10\% of the total number of GPUs, and 62\% had access to less then 8 GPUs.
Only a very small faction of researchers (2.2\% of our respondents) had access to GPU compute (1000 or more) to train models with several hundreds of billion parameters for several days or weeks.
Many researchers, hence, could only fine-tune models---which requires far fewer resources than pre-training~\citep{zhou-etal-2021-hulk}---which is only possible when pre-trained model weights are available. Unfortunately, many recent models are being kept private, which has intensified the discussion about equity in the field~\citep{togelius2023choose}. 

\section{Impact on Reviewing}\label{sec:reviewing}

\begin{figure*}[!htb]
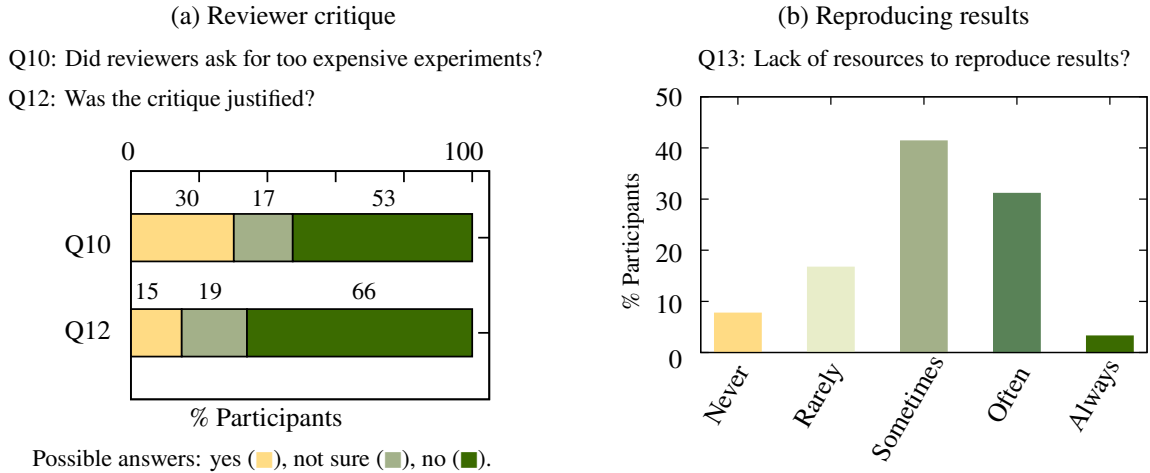

    \begin{subfigure}[b]{0.49\textwidth}
        \centering
        \caption{Reviewer critique} 
        \label{fig:q10-12}
        \include{final_plots/q10_12_bars}
        %\vspace{-2.5em}
    \end{subfigure}
    \hfill
    \begin{subfigure}[b]{0.49\textwidth}
        \centering
        \caption{Reproducing results} 
        \label{fig:q13}
        \include{final_plots/q13_bars}
    \end{subfigure}
    \caption{Analysis on how of a lack of resources can affect research. In (a), we show what percentage of participants had been asked by reviewers for too expensive experiments ($Q10$) and if so, if they felt the critique was justified ($Q12$). In (b), we show how often our participants could not reproduce previous results due to a lack of computational resources ($Q13$).}
    \label{fig:q10-13}
\end{figure*}

Finally, we quantified how the concerns about the environmental impact and differences in terms of available compute resources affect peer reviewing ($Q$10--$Q$13). 
We further asked our participants four questions ($Q$14--$Q$17) which relate to concrete ideas that would change the reviewing process and encourage model release ($Q18$).

\subsection{Analysis}
\Cref{fig:q10-12} shows that 30.1\% of the participants experienced being asked (during peer-review) to conduct additional experiments that were too expensive for them ($Q$10) with 77 respondents having experienced this more than once ($Q11$) and 19.2\% having a substantially higher number (five or more times) according to our outlier analysis.
Most participants (65.9\%) further thought that this criticism was unjustified ($Q$12). 
\Cref{fig:q13} ($Q$13) shows that 34.3\% of the respondents \textit{often} or \textit{always} lacked resources to reproduce previous experiments and 41.4\% \textit{sometimes}. 
Only 7.7\% \textit{never} or 16.7\% \textit{rarely} faced a lack of resources to reproduce experiments.

\begin{figure*}[!htb]
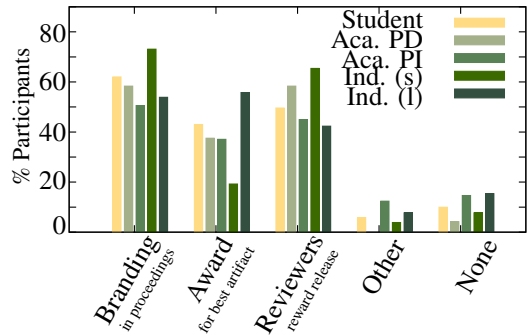

    \begin{subfigure}[b]{0.44\textwidth}
        \centering
        \caption{Potential changes to reviewing}\label{fig:q14-17}
        \vspace{3em}
        \include{final_plots/q14_17_bars}
    \end{subfigure}
    \hfill 
    \begin{subfigure}[b]{0.44\textwidth}
        \centering
        \caption{Model release}\label{fig:q18}
        \include{final_plots/q18_bars}
    \end{subfigure}

    \begin{subfigure}[b]{0.44\textwidth}
        \centering
        \vspace{2em}
        \caption{Model release (by seniority)} 
        \label{fig:q18-age}
        \vspace{-1.5em}
        \include{final_plots/q18_age}
    \end{subfigure}
    \hfill 
    \begin{subfigure}[b]{0.44\textwidth}
        \centering
        \caption{Model release (by job sector)} 
        \label{fig:q18-job}
        \vspace{-1.5em}
        \include{final_plots/q18_job}
    \end{subfigure}
    \caption{Analysis of responses on how to improve the reviewing process. In (a), we show the distribution of our participants' responses for $Q14$--$Q17$ (in \%). A majority of our participants would submit to an efficiency track ($Q14$) and would prefer reviewers to justify a request for more experiments ($Q16)$. They further would benefit from a release of smaller models ($Q16$). In contrast, the responses are more mixed about the authors justifying the compute budget ($Q15$). In (b--d), we show our participants' responses on how to encourage the release of models (in \%): (b) overall, (c) by seniority, (d) by job sector. Multiple responses were allowed for $Q18$.}
    \label{fig:q18-fine-grained}
\end{figure*}

With respect to the concrete reviewing actions, \cref{fig:q14-17} shows that a large majority (89.8\%) of our participants would consider submitting their work to a dedicated track on efficient methods ($Q$14).
Following up on the results from the survey, such an efficiency track was implemented at EMNLP 2022.
35.9\% of our participants were unsure about requesting authors to justify the allocation of budget for experiments ($Q$15), with 41\% voting for \textit{yes}.
Also, even though 52.6\% of the participants had not been asked for experiments that were too expensive for them, a clear majority of the participants (83.7\%) would like to require reviewers to justify their petitions for more experiments ($Q$16).
Lastly, we also see a large majority (75.6\%) that believed that their work could benefit from the release of small versions of pretrained models alongside large ones ($Q$17). 
To promote this, a majority of our respondents thought that venues should have a visible \textit{branding} of papers to release a model (59.3\%) and that reviewers should be instructed to reward model release (50.6\%).
42.6\% of respondents thought that the venues should grant a best artifact \textit{award}. 11.5\% of respondents supported \textit{none} of the options.
A first step towards increasing the reproducibility and ensuring the submission of experimental code was implemented at NAACL 2022 by introducing a badge system at the reproducibility track.\footnote{\url{https://2022.naacl.org/blog/reproducibility-track/}}
Upon acceptance, the authors could follow specific procedures to earn three types of badges: 1) open-source code, 2) trained model, and 3) reproducible results.

\paragraph{Seniority.}
We find no significant differences w.r.t.~the seniority of our participants regarding $Q10$--$Q18$. However, junior researchers (1--5 years) showed a substantially higher tendency towards requesting authors to justify their compute budget ($Q$15) against all other age groups (p-values $<0.035$). 
We also observe in \Cref{fig:q18-age} diverging preferences between junior and senior groups in terms of ideas to improve the reviewing process ($Q$18). Junior researchers (1--5 years) seemed to be more inclined towards a visual branding as well as instructing reviewers than senior researchers (11--15 years with a p-value of $0.089$ and 16+ years with a p-value of $0.085$).

\begin{figure}[t]
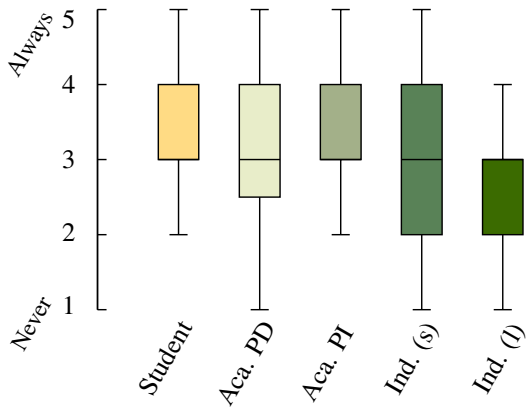
{}
    \include{final_plots/q13_job}
    \caption{$Q$13: Lack of resources by job sector.} 
    \label{fig:q13-job}
\end{figure}

\paragraph{Job Sector.}
In terms of the job sector, we again find no significant differences with respect to reviewers asking for too expensive experiments ($Q$10) or critique being justified ($Q$12).
Interestingly, respondents from small industry received fewer such requests ($Q11$) compared to post-docs (p-value $=0.024$), PIs (p-value $= 0.061$), and larger industry (p-value $=0.087$). 
The most concerning trend can be observed when comparing the different groups with respect to their lack of compute resources to reproduce experiments ($Q$13, \cref{fig:q13-job}); where we find significant differences and conduct pairwise analyses.\footnote{Kruskal-Wallis test: $H^5 =12.486 > H_0^5=9.488$.}
In general, students suffered most, with a significant difference compared to the large industry sector with a p-value of $0.002 < 0.005 = \alpha$ (Bonferroni-corrected).  
We further find substantial differences between students and academic PIs (p-value $=0.026$) and between academic post-docs and large industry labs (p-value $=0.088$).

We find no substantial differences when it comes to actionable items for the *CL community ($Q$14--$Q$17), indicating that implementing popular ideas would be welcomed by all groups.
However, we find some differences when it comes to encouraging the release of models ($Q$18).
For instance, \Cref{fig:q18-job} shows that academic post-docs had a higher preference towards reviewers rewarding papers that promise to release models than academic PIs. Also, participants from small industry would prefer visual branding over awards in contrast to large industry.

\subsection{Discussion and Recommendations}\label{sec:reviewing-discussion}
Our analysis shows that the two most pressing issues among our respondents are the lack of resources to reproduce results and reviewers requesting for too expensive experiments without proper justification.
This is reflected in the large support for both respective counter measures; namely, asking reviewers to provide justification and the release of smaller models that would allow researchers to at least reproduce some experiments.
Considering the success of badges at NAACL 2022 with 175 code, 98 model and 20 reproducibility badges, introducing an explicit badge for small model release could boost inclusiveness and reproducibility.\footnote{\url{https://naacl2022-reproducibility-track.github.io/results/}} 
To improve peer reviewing, one immediate action could be to adapt the ARR reviewing guidelines and instruct reviewers to consider the compute budget reported in a paper when asking for more experiments.\footnote{\url{https://aclrollingreview.org/reviewertutorial}} %; for instance by considering experiments that are within the reported compute budget of the paper.
%For guidance, this could be accompanied by an example showing how to consider the available compute power reported in a paper. %in \textit{check for lazy thinking} (section 6).

Among the 22 additional suggestions for $Q18$, we find a high emphasis (68.2\%) towards the release of artifacts---both because this facilitates future research and helps reproducibility.
Moreover, 22 of the 67 general suggestion ($Q19$) also touched upon issues about model release and reviewing, highlighting the importance of both topics.
The responses mentioned a remarkably wide variety of artifacts: code; trained models; system outputs (to facilitate comparative evaluations without rerunning the code); training checkpoints (to study the training dynamics); and proper documentation of training data (including crowdsourcing questions).
In addition to simply releasing trained models, several respondents also wished for a sufficiently high quality of the released models complemented by code and documentation. 
One particular concern was how the release of artifacts should be integrated into the reviewing process. 
On the one hand, it seems useful to submit artifacts together with the paper before reviewing, so that reviewers can access them and to prevent breaking promises of future code release. 
On the other hand, this needs to happen within the constraints of double-blind reviewing.
Finally, 12 of the free-text responses of $Q18$ and $Q19$ suggested that artifact release should be mandatory for acceptance.

\section{Further Considerations}
Finally, we discuss suggestions ($Q19$) that do not fit into any of the previously discussed topics.
From the 67 free-text responses (21.5\%), the two most prominent topics were evaluation (11 respondents) and emphasizing research over engineering (7 respondents).

\paragraph{Evaluation.} 
16.4\% of the free-text respondents touched upon the issue of evaluation and model comparability; as current benchmarks often focus on improving a single metric.
One measure to counter this trend would be to report performance based on Pareto frontiers and to consider the compute budget along with the model performance.
To promote such curves, it would also be important to release metadata including preprocessing and hyperparameter choices that allows future research to draw proper comparisons as well as to provide concrete guidelines for reviewers.

\paragraph{Research vs.~engineering.} 
10.4\% of the free-text respondents further noted that the field seemed to have drifted more towards engineering by primarily chasing high performance; straying away from producing meaningful scientific insights.
The respondents brought forward various suggestions to combat this; for instance that authors should clearly state their scientific hypothesis and then report research that tests this hypothesis using the lowest appropriate amount of resources.
Other suggestions were to actively promote more theoretical, or more non deep learning work.

\paragraph{Other suggestions.}
Another suggestion worth mentioning was the creation of a separate track (four respondents); either specifically for small models or for industry that cannot publish their models.
Finally, there was also a call for more shared tasks with limited resources such as the efficient NMT challenge~\citep{heafield-etal-2022-findings} or the efficient inference task~\citep{sustainlp-2020-sustainlp}.

\section{Conclusion}
We presented a first attempt to capture and quantify existing concerns about the environmental impact and equity within the *CL community.
We further investigated the resulting implications on peer reviewing considering the increasing computational demand.
A majority of our respondents were concerned regarding the environmental footprint of NLP experiments with model training and model selection being the most pressing issues.
We also found a high disparity among our respondents with students and small industry researchers suffering most from a lack of resources.
There was a large support for measures to improve equity and accessibility across all respondents; most prominently for an efficiency track, asking reviewers to justify the petition for additional experiments, and the release of small versions of pretrained models.

Considering the continuous increase of parameters in PLMs~\citep{zhao2023survey}, one danger we face is that existing disparities may intensify even further. 
However, we find that much can be done to combat this, even on an individual level.
As a researcher, by making our model weights, code, and data publicly available; and as a reviewer, by being considerate towards the available compute budget.

\section*{Limitations}
To receive a large number of responses, this survey was advertised throughout various channels.
Hence, this is by no means a representative study within the whole *CL community. 
This is partially reflected in the evaluation of the geographic locations, e.g., they were too coarse to capture a more precise picture about existing geographic inequalities.  
Nonetheless, the fact that we did not receive any responses from bodies located in Africa indicates that there may exist high disparities in terms of geographic location. 
For the same reason, the disparities found in this survey are more indicative than representative.
Consequently, any action that is being implemented should not be solely derived from the survey data and carefully considered beforehand.

\section*{Acknowledgements}
This work was initiated at and benefited substantially from the Dagstuhl Seminar 22232: \textit{Efficient and Equitable Natural Language Processing in the Age of Deep Learning}.
%TODO: add other participants here
We further thank Niranjan Balasubramanian, Jonathan Frankle, Michael Hassid, Kenneth Heafield, Sara Hooker, Alexander Koller, Alexandra Sasha Luccioni,  Alexander L{\"o}ser, André F. T. Martins, Colin Raffel, Nils Reimers, Leonardo Riberio, Anna Rogers, Edwin Simpson, Noam Slonim, Noah A. Smith, and Thomas Wolf for a fruitful discussion and helpful feedback at the seminar. 
We further thank Leshem Choshen for helpful feedback on this work.

\bibliography{custom}

\begin{thebibliography}{20}
\expandafter\ifx\csname natexlab\endcsname\relax\def\natexlab#1{#1}\fi

\bibitem[{Abramowitz(1974)}]{abramowitz1988handbook}
Milton Abramowitz. 1974.
\newblock \emph{Handbook of Mathematical Functions, With Formulas, Graphs, and
  Mathematical Tables,}.
\newblock Dover Publications, Inc., USA.

\bibitem[{Arase et~al.(2021)Arase, Blunsom, Diab, Dodge, Gurevych, Liang,
  Raffel, R{\"{u}}ckl{\'{e}}, Schwartz, Smith, Strubell, and
  Zhang}]{ACL-policy-doc-2021}
Yuki Arase, Phil Blunsom, Mona Diab, Jesse Dodge, Iryna Gurevych, Percy Liang,
  Colin Raffel, Andreas R{\"{u}}ckl{\'{e}}, Roy Schwartz, Noah~A. Smith, Emma
  Strubell, and Yue Zhang. 2021.
\newblock \href
  {https://www.aclweb.org/portal/content/efficient-nlp-policy-document}
  {{Efficient NLP policy document}}.

\bibitem[{Bonferroni(1936)}]{bonferroni1936teoria}
Carlo Bonferroni. 1936.
\newblock \href {https://doi.org/http://dx.doi.org/10.4135/9781412961288.n455}
  {Teoria statistica delle classi e calcolo delle probabilita}.
\newblock \emph{Pubblicazioni del R Istituto Superiore di Scienze Economiche e
  Commericiali di Firenze}, 8:3--62.

\bibitem[{Fisher(1921)}]{fisher1921probable}
Roland~A. Fisher. 1921.
\newblock On the" probable error" of a coefficient of correlation deduced from
  a small sample.
\newblock \emph{Metron}, 1:3--32.

\bibitem[{Heafield et~al.(2022)Heafield, Zhang, Nail, Van Der~Linde, and
  Bogoychev}]{heafield-etal-2022-findings}
Kenneth Heafield, Biao Zhang, Graeme Nail, Jelmer Van Der~Linde, and Nikolay
  Bogoychev. 2022.
\newblock \href {https://aclanthology.org/2022.wmt-1.4} {Findings of the {WMT}
  2022 shared task on efficient translation}.
\newblock In \emph{Proceedings of the Seventh Conference on Machine Translation
  (WMT)}, pages 100--108, Abu Dhabi, United Arab Emirates (Hybrid). Association
  for Computational Linguistics.

\bibitem[{Izsak et~al.(2021)Izsak, Berchansky, and Levy}]{izsak2021}
Peter Izsak, Moshe Berchansky, and Omer Levy. 2021.
\newblock \href {https://doi.org/10.18653/v1/2021.emnlp-main.831} {How to train
  {BERT} with an academic budget}.
\newblock In \emph{Proceedings of the 2021 Conference on Empirical Methods in
  Natural Language Processing}, pages 10644--10652, Online and Punta Cana,
  Dominican Republic. Association for Computational Linguistics.

\bibitem[{Kruskal and Wallis(1952)}]{kruskal1952use}
William~H. Kruskal and W.~Allen Wallis. 1952.
\newblock \href {http://www.jstor.org/stable/2280779} {{Use of Ranks in
  One-Criterion Variance Analysis}}.
\newblock \emph{Journal of the American Statistical Association},
  47(260):583--621.

\bibitem[{Levene(1960)}]{levene1960robust}
Howard Levene. 1960.
\newblock Robust tests for equality of variances.
\newblock \emph{Contributions to probability and statistics}, pages 278--292.

\bibitem[{Moosavi et~al.(2020)Moosavi, Fan, Shwartz, Glava{\v{s}}, Joty, Wang,
  and Wolf}]{sustainlp-2020-sustainlp}
Nafise~Sadat Moosavi, Angela Fan, Vered Shwartz, Goran Glava{\v{s}}, Shafiq
  Joty, Alex Wang, and Thomas Wolf, editors. 2020.
\newblock \href {https://aclanthology.org/2020.sustainlp-1.0}
  {\emph{Proceedings of SustaiNLP: Workshop on Simple and Efficient Natural
  Language Processing}}. Association for Computational Linguistics, Online.

\bibitem[{Patterson et~al.(2022)Patterson, Gonzalez, H{\"o}lzle, Le, Liang,
  Munguia, Rothchild, So, Texier, and Dean}]{patterson2022carbon}
David Patterson, Joseph Gonzalez, Urs H{\"o}lzle, Quoc Le, Chen Liang,
  Lluis-Miquel Munguia, Daniel Rothchild, David~R So, Maud Texier, and Jeff
  Dean. 2022.
\newblock The carbon footprint of machine learning training will plateau, then
  shrink.
\newblock \emph{Computer}, 55(7):18--28.

\bibitem[{Schwartz et~al.(2020)Schwartz, Dodge, Smith, and
  Etzioni}]{schwartz2020green}
Roy Schwartz, Jesse Dodge, Noah~A Smith, and Oren Etzioni. 2020.
\newblock Green ai.
\newblock \emph{Communications of the ACM}, 63(12):54--63.

\bibitem[{Shaphiro and Wilk(1965)}]{shaphiro1965analysis}
S~Shaphiro and MBJB Wilk. 1965.
\newblock An analysis of variance test for normality.
\newblock \emph{Biometrika}, 52(3):591--611.

\bibitem[{Strubell et~al.(2019)Strubell, Ganesh, and
  McCallum}]{strubell-etal-2019-energy}
Emma Strubell, Ananya Ganesh, and Andrew McCallum. 2019.
\newblock \href {https://doi.org/10.18653/v1/P19-1355} {Energy and policy
  considerations for deep learning in {NLP}}.
\newblock In \emph{Proceedings of the 57th Annual Meeting of the Association
  for Computational Linguistics}, pages 3645--3650, Florence, Italy.
  Association for Computational Linguistics.

\bibitem[{Togelius and Yannakakis(2023)}]{togelius2023choose}
Julian Togelius and Georgios~N Yannakakis. 2023.
\newblock Choose your weapon: Survival strategies for depressed ai academics.
\newblock \emph{arXiv preprint arXiv:2304.06035}.

\bibitem[{Treviso et~al.(2023)Treviso, Lee, Ji, Aken, Cao, Ciosici, Hassid,
  Heafield, Hooker, Raffel, Martins, Martins, Forde, Milder, Simpson, Slonim,
  Dodge, Strubell, Balasubramanian, Derczynski, Gurevych, and
  Schwartz}]{treviso2022efficient}
Marcos Treviso, Ji-Ung Lee, Tianchu Ji, Betty~van Aken, Qingqing Cao, Manuel~R.
  Ciosici, Michael Hassid, Kenneth Heafield, Sara Hooker, Colin Raffel,
  Pedro~H. Martins, André F.~T. Martins, Jessica~Zosa Forde, Peter Milder,
  Edwin Simpson, Noam Slonim, Jesse Dodge, Emma Strubell, Niranjan
  Balasubramanian, Leon Derczynski, Iryna Gurevych, and Roy Schwartz. 2023.
\newblock \href {https://doi.org/10.1162/tacl_a_00577} {{Efficient Methods for
  Natural Language Processing: A Survey}}.
\newblock \emph{Transactions of the Association for Computational Linguistics},
  11:826--860.

\bibitem[{Tukey(1977)}]{tukey77}
John~W. Tukey. 1977.
\newblock \emph{Exploratory Data Analysis}.
\newblock Addison-Wesley.

\bibitem[{Welch(1951)}]{welch1951comparison}
Bernard~Lewis Welch. 1951.
\newblock \href {https://doi.org/https://doi.org/10.1093/biomet/38.3-4.330}
  {{On the Comparison of Several Mean Values: An Alternative Approach}}.
\newblock \emph{Biometrika}, 38(3/4):330--336.

\bibitem[{Wu et~al.(2022)Wu, Raghavendra, Gupta, Acun, Ardalani, Maeng, Chang,
  Aga, Huang, Bai et~al.}]{wu2022sustainable}
Carole-Jean Wu, Ramya Raghavendra, Udit Gupta, Bilge Acun, Newsha Ardalani,
  Kiwan Maeng, Gloria Chang, Fiona Aga, Jinshi Huang, Charles Bai, et~al. 2022.
\newblock Sustainable ai: Environmental implications, challenges and
  opportunities.
\newblock \emph{Proceedings of Machine Learning and Systems}, 4:795--813.

\bibitem[{Zhao et~al.(2023)Zhao, Zhou, Li, Tang, Wang, Hou, Min, Zhang, Zhang,
  Dong et~al.}]{zhao2023survey}
Wayne~Xin Zhao, Kun Zhou, Junyi Li, Tianyi Tang, Xiaolei Wang, Yupeng Hou,
  Yingqian Min, Beichen Zhang, Junjie Zhang, Zican Dong, et~al. 2023.
\newblock A survey of large language models.
\newblock \emph{arXiv preprint arXiv:2303.18223}.

\bibitem[{Zhou et~al.(2021)Zhou, Chen, Jin, and Wang}]{zhou-etal-2021-hulk}
Xiyou Zhou, Zhiyu Chen, Xiaoyong Jin, and William~Yang Wang. 2021.
\newblock \href {https://doi.org/10.18653/v1/2021.eacl-demos.39} {{HULK}: An
  energy efficiency benchmark platform for responsible natural language
  processing}.
\newblock In \emph{Proceedings of the 16th Conference of the European Chapter
  of the Association for Computational Linguistics: System Demonstrations},
  pages 329--336, Online. Association for Computational Linguistics.

\end{thebibliography}
\bibliographystyle{acl_natbib}

\end{document}